\documentclass[10pt,twocolumn,letterpaper]{article}

\usepackage[pagenumbers]{cvpr} 

\usepackage[ruled,linesnumbered]{algorithm2e}
\usepackage{multicol}
\usepackage{multirow}
\usepackage[table]{xcolor}

\definecolor{cvprblue}{rgb}{0.21,0.49,0.74}
\usepackage[utf8]{inputenc}
\usepackage[pagebackref,breaklinks,colorlinks,allcolors=cvprblue]{hyperref}

\title{SpotEdit: Selective Region Editing in Diffusion Transformers}

\author{
Zhibin Qin$^{1}$\footnotemark[1]\quad
Zhenxiong Tan$^{1}$\footnotemark[1] \quad
Zeqing Wang$^{1}$ \quad
Songhua Liu$^{2}$ \quad
Xinchao Wang$^{1}${\footnotemark[2]}
\\
$^{1}$National University of Singapore \quad
$^{2}$Shanghai Jiao Tong University
\\
{\tt\small
\{e1352224, zhenxiong, zeqing.wang\}@u.nus.edu \quad
liusonghua@sjtu.edu.cn \quad
xinchao@nus.edu.sg
}
}
\begin{document}

\twocolumn[{
\maketitle
\begin{center}
\vspace{-1.6em}
    \centering
    \captionsetup{type=figure}
    \includegraphics[width=\textwidth]{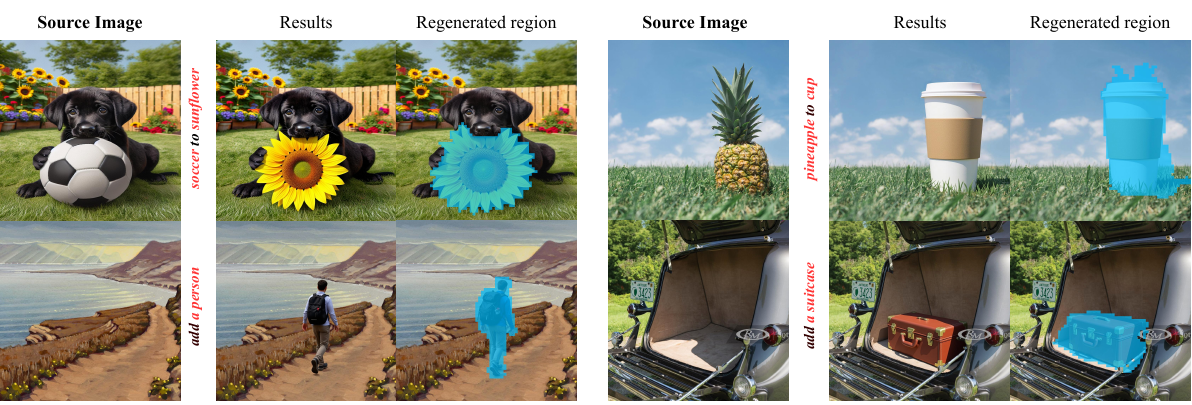}
    \caption{Examples of edited images by SpotEdit. The blue area reveals the regenerated region. }
    \label{fig:Qualitative_Results}
\end{center} 
}]

{
\renewcommand{\thefootnote}{\fnsymbol{footnote}} 
\footnotetext[1]{Eqaul contribution.}
\footnotetext[2]{Corresponding author.}
}
\begin{abstract}

Diffusion Transformer models have significantly advanced image editing by encoding conditional images and integrating them into transformer layers.
However, most edits involve modifying only small regions, while current methods uniformly process and denoise all tokens at every timestep, causing redundant computation and potentially degrading unchanged areas.
This raises a fundamental question: \emph{Is it truly necessary to regenerate every region during editing?}
To address this, we propose \textit{SpotEdit}, a training-free diffusion editing framework that selectively updates only the modified regions.
SpotEdit comprises two key components: \textit{SpotSelector} identifies stable regions via perceptual similarity and skips their computation by reusing conditional image features; \textit{SpotFusion} adaptively blends these features with edited tokens through a dynamic fusion mechanism, preserving contextual coherence and editing quality.
By reducing unnecessary computation and maintaining high fidelity in unmodified areas, SpotEdit achieves efficient and precise image editing.
Project page is available at \url{https://biangbiang0321.github.io/SpotEdit.github.io/}.

\end{abstract}
\newcommand{\ourmtd}[0]{~{SpotEdit}~}

\section{Introduction}
\label{sec:intro}

\begin{figure*}[htbp]
    \centering
    \includegraphics[width=0.9\textwidth]{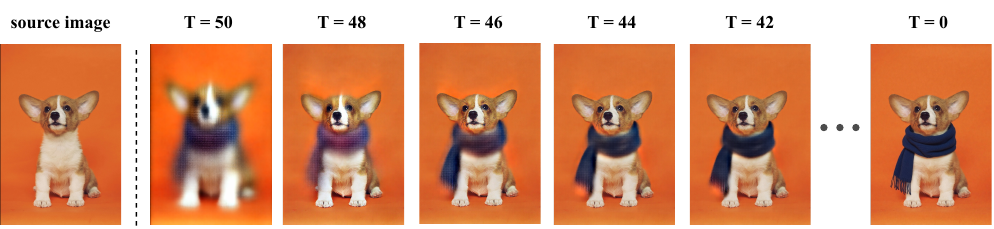}
    \caption{ Reconstruction results at different timesteps.(Totalsteps T=50, seed=42, prompt = ``Add a scarf to the dog.'') Each reconstruction $\hat{x_0}$ is estimated following the rectified flow formulation:
$\hat{X_0} = x_t - t \, v_{\theta}(x_t, c, t)$. It can be observed that some regions become sharp and visually consistent with the original image even at very early stages, while other regions continue to evolve until the final timestep.}
    \label{fig:recon_sequence}
\end{figure*}
Diffusion models have shown outstanding performance in image generation tasks~\cite{hoClassifierFreeDiffusionGuidance2022,rombach2021highresolution, labs2025flux}.
By leveraging Diffusion Transformer (DiTs) architectures~\cite{peeblesScalableDiffusionModels2023}, the generation quality and flexibility have been further enhanced.
Building on these advances, encoding condition images and integrating them directly into transformers~\cite{tan2025ominicontrol, tan2025ominicontrol2} has become a mainstream technique for image editing~\cite{labsFLUX1KontextFlow2025,wuQwenImageTechnicalReport2025}.
This strategy allows simple yet effective editing without relying on manually provided masks, greatly improving usability in practical applications.

However, in most image editing tasks, only a small region of the image requires modification, while the majority of areas remain unchanged. 
Yet, existing approaches uniformly follow a full-image regeneration paradigm, indiscriminately denoising every region from random noise, including those that do not need editing. 
Such uniform processing introduces two prominent drawbacks: First, redundant computations in non-edited regions may inadvertently produce subtle artifacts; second, significant computational resources are wasted by processing unmodified areas.
These issues naturally lead us to reconsider the current editing paradigm and pose a critical question: \textbf{\textit{Is it truly necessary to regenerate every region of the image during an editing task?}}



To address the above problems, we begin by analyzing the temporal convergence patterns of latent representations during diffusion.
Figure~\ref{fig:recon_sequence} reveal that, in partial editing tasks, non-edited regions stabilize quickly, converging at early diffusion timesteps.
This observation naturally motivates a more efficient editing strategy: \emph{Edit only what needs to be edited}.

Guided by this principle, we propose \ourmtd, a mechanism designed to automatically detect stable, non-edited regions and reuse their corresponding condition image latent features without computing them in DiTs, thereby avoiding redundant regeneration computation.
Implementing this idea, however, raises two critical challenges: 
1) How to efficiently and accurately identify non-edited regions?
2) How to enable models to dynamically focus computation only on the regions requiring modification?

For challenge 1, we propose SpotSelector, an adaptive mechanism that dynamically identifies stable regions during diffusion iterations. 
Specifically, SpotSelector computes a perceptual similarity score for each latent token by measuring the perceptual distance between the reconstructed fully denoised latent and the corresponding condition image latent via VAE decoder layers.
Regions whose perceptual distance is below a threshold are automatically classified as non-edited regions. This approach eliminates manual masking and directly leverages the diffusion dynamics observed in our analysis, ensuring that the identified regions align with the model's generative process.

For challenge 2, we introduce \textit{SpotFusion}, a context fusion mechanism that restores missing contextual information by adaptively blending features from the condition image. Leveraging the high feature relationship between non-edited regions and corresponding condition image regions across diffusion steps, SpotFusion dynamically modulates the contribution of the reference based on the current denoising timestep, relying more on the reference early in the process and gradually shifting to the current estimate as generation progresses. This design preserves time coherence in features while avoiding potential boundary artifacts, without requiring additional computation for unedited regions.

Experimental results demonstrate that our SpotEdit achieves a speedup of $1.7\times$ for imgEdit-Benchmark\cite{ye2025imgEdit}and $1.9\times$ for PIE-Bench++\cite{huangParallelEditsEfficientMultiobject} on the base model FLUX.1-Kontext\cite{labsFLUX1KontextFlow2025}, while maintaining quality comparable to the original model. Qualitative results (see Figure~\ref{fig:Qualitative_Results}) further indicate that SpotEdit perfectly preserves non-edited regions and produces clean, localized edits.


Our primary contributions are summarized as follows:
\begin{enumerate}[label=(\roman*)]
    \item We propose SpotSelector, a perceptual-similarity-based method for dynamically distinguishing non-edited regions, removing the need for manual masks.
    \item We introduce SpotFusion, an adaptive fusion mechanism ensuring temporal coherence and contextual consistency in partially-edited diffusion processes.
    \item We demonstrate that our combined framework, SpotEdit, enables selective diffusion-based editing, significantly accelerating inference while preserving the fidelity and quality of edits.
\end{enumerate}

\section{Related works}
\label{sec:related_works}

\subsection{Precise image editing}
Image editing has long been a central need in workflows. \cite{Prez2003PoissonIE}Following the remarkable breakthroughs of diffusion models~\cite{ho2020denoisingdiffusionprobabilisticmodels,yang2025diffusionmodelscomprehensivesurvey} in image generation, a growing body of research has focused on adapting these models to serve image editing tasks. Early approaches, such as ControlNet~\cite{zhang2023adding}, injected external control signals into U-Net~\cite{ronneberger2015unetconvolutionalnetworksbiomedical} to enable robust and controllable editing outcomes.

With the advancement of diffusion models, inversion-based methods~\cite{zhuKVEditTrainingFreeImage2025,yanEEditRethinkingSpatial2025,hertzPrompttoPromptImageEditing2022,longFollowYourShapeShapeAwareImage2025,wangTamingRectifiedFlow2025,meng2022sdeditguidedimagesynthesis,hertz2022prompttopromptimageeditingcross} have become the mainstream paradigm.
These approaches operate by injecting noise into the condition image and then denoising it under the target textual prompt to produce the edited result\cite{Wu2024TurboEditIT}. However, the pure noise-addition and denoising procedure often leads to unwanted global changes in the image.

To better adhere to localized editing requirements, recent works introduce KV-injection techniques~\cite{zhuKVEditTrainingFreeImage2025,longFollowYourShapeShapeAwareImage2025,hertzPrompttoPromptImageEditing2022,kawarImagicTextBasedReal2023,mokady2022nulltextinversioneditingreal,wangTamingRectifiedFlow2025,Avrahami2021BlendedDF} that preserve reference features throughout the denoising process.
Inversion KV is injected into the matching timesteps of the denoising phase, effectively transferring the structural and semantic information from the condition image to the edited output~\cite{zhuKVEditTrainingFreeImage2025,longFollowYourShapeShapeAwareImage2025,wangTamingRectifiedFlow2025,hertz2022prompttopromptimageeditingcross}.

In parallel, another line of work adopts mask-based control mechanisms~\cite{skaik2024mcgmmaskconditionaltexttoimage,Sheynin2023EmuEP}.
These approaches solve the editing task by inpainting and explicitly specify editable regions via a binary mask\cite{Couairon2022DiffEditDS}, allowing the denoising process to focus computations only within the masked area while directly retaining unedited pixels from the condition image~\cite{zhuKVEditTrainingFreeImage2025,tan2025ominicontrol,tan2025ominicontrol2,chang2022maskgitmaskedgenerativeimage,zhangMagicBrushManuallyAnnotated2024}. Such strategies offer fine-grained control over editing region; however, they limits flexibility and applicability in real-world scenarios.

More recently, a new generation of image editing models~\cite{labsFLUX1KontextFlow2025,wuQwenImageTechnicalReport2025} have emerged, leveraging Diffusion Transformer architectures~\cite{peeblesScalableDiffusionModels2023} to jointly process condition images and noise inputs. This design requires only high-level editing instructions and can directly modify the corresponding semantic regions~\cite{brooks2023instructpix2pixlearningfollowimage,Parmar2023ZeroshotIT,hertz2022prompttopromptimageeditingcross}, without relying on any manually provided masks.

Despite the rapid progress of precise image editing, all the mentioned approaches share a fundamental limitation: they regenerate the entire image at every denoising timestep, regardless of which regions actually require modification.
\subsection{Acceleration of efficient editing}

Existing acceleration techniques for diffusion and diffusion–transformer models mainly operate at the full-token level\cite{lyu2022acceleratingdiffusionmodelsearly}, without distinguishing between edited and non-edited regions.
Acceleration methods~\cite{selvarajuFORAFastForwardCaching2024,chen$D$DiTTrainingFreeAcceleration2024,liuTimestepEmbeddingTells2025,yuan2024ditfastattn,liuReusingForecastingAccelerating2025,chuOmniCacheTrajectoryOrientedGlobal2025,ma2024learningtocacheacceleratingdiffusiontransformer,chen2025regione} improve efficiency by reusing or approximating intermediate features across timesteps.
While effective for generation, these approaches treat all image tokens uniformly, accelerating every spatial position regardless of its semantic importance.
However, image editing tasks naturally contain heterogeneous regions: only a small subset of tokens needs modification, whereas the majority should remain unchanged.
Quality degradation introduced by aggressive full-token acceleration tends to disproportionately affect the semantically important edited regions, leading to visibly inferior results even when background distortion is mild.
Consequently, full-token acceleration provides limited practical benefit for editing, as the gain in speed often comes with a noticeable loss in fidelity.

On the other hand, recent token-space strategies such as ToCa~\cite{zouAcceleratingDiffusionTransformers2024}, DUCA~\cite{zou2024DuCa}, and RAS~\cite{liuRegionAdaptiveSamplingDiffusion2025} explore reducing computational redundancy by compacting tokens or adjusting sampling steps based on token saliency.
Although these approaches operate at the token level, they still follow the paradigm of reducing timesteps or compressing token representations, rather than explicitly skipping computation for non-edited regions.
As a result, they do not fully exploit the inherent sparsity of image editing tasks and cannot direct computational resources to edited regions alone.
Thus, none of the above methods achieve true region-aware acceleration, where the model selectively updates only the edited tokens while bypassing unnecessary computation for preserved regions.


\section{Preliminary}
\label{sec:prelim_flow}
Flow Matching~\cite{lipman2023flowmatchinggenerativemodeling} formulates generative modeling as learning a deterministic flow that continuously transports samples from a source distribution \( p_1 \), typically Gaussian noise, to a target distribution \( p_0 \) representing real images.  
A time-dependent velocity field \( v_\theta(X, C, t) \) defines transformation through the ordinary differential equation
\begin{equation}
\frac{dX_t}{dt} = v_\theta(X_t, C, t),
\quad X_1 \sim p_1, \; X_0 \sim p_0,
\label{eq:flow_ode}
\end{equation}
where \( C \) denotes the condition that contains the reference image and the instruction describing the desired edit.  
During inference, the model generates samples by integrating Eq.~\ref{eq:flow_ode} backward in time from \( X_1 \) to \( X_0 \).

Rectified Flow~\cite{liu2022flowstraightfastlearning} simplifies this formulation by assuming a linear interpolation between the source and target;
\begin{equation}
X_t = (1 - t) X_0 + t X_1,
\quad v_\theta(X_t, C, t) = X_1 - X_0
\label{eq:rectified_flow}
\end{equation}
with endpoints sampled as \( X_0 \sim p_0 \) drawn from the data distribution and \( X_1 \sim p_1 \) following \( \mathcal{N}(0, I) \).  
On this linear path, the instantaneous target velocity becomes \( u_t \equiv X_1 - X_0 \), resulting in a constant and easily learnable flow field.

For inference, the reverse process is approximated.
\begin{equation}
X_{t_{i-1}} = X_{t_i} - (t_i - t_{i-1}) \, v_\theta(X_{t_i}, C, t_i),
\quad i = T, \dots, 1
\label{eq:inference}
\end{equation}
which progressively transports the initial noise sample \( X_1 \) to the final image \( X_0 \) under the guidance of condition \( C \).

\section{Methodology}
\label{sec:methodology}

\begin{figure*}[htbp]
    \centering
    \includegraphics[width=0.9\textwidth]{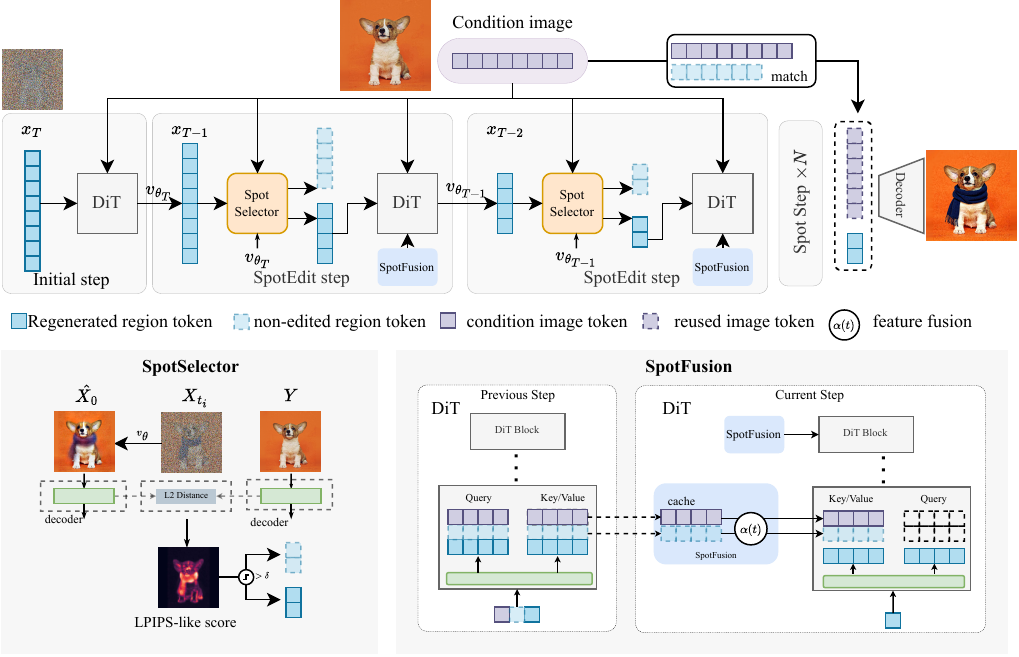}
    \caption{\textbf{Overview of SpotEdit.} 
    The process consists of three stages: 
    (1) \textbf{Initial Steps}: the model performs standard DiT denoising on all image tokens under the editing instruction, while caching the KV values for Spotfusion.  
    (2) \textbf{Spot Steps}: SpotSelector \emph{dynamically} identifies regenerated region and non-edited region tokens using LPIPS-like perceptual scores. Non-edited region tokens are skipped by DiT computation, while regenerated region tokens are generated iteratively with SpotFusion, which builds a temporally consistent condition cache by fusing cached non-edited region KV values with condition image KV values.
    (3) \textbf{Token Replacement}: Regenerated tokens are updated through DiT, and non-edited tokens are directly covered by the corresponding reused tokens before decoding into an image, ensuring background fidelity with reduced computation.}
    \label{fig:pipeline}
\end{figure*}

As discussed in Section~\ref{sec:intro},  current mainstream instruction-based editing models(FLUX-Kontext\cite{labsFLUX1KontextFlow2025}, Qwen-Image-Editing\cite{wuQwenImageTechnicalReport2025}) regenerate the entire image during the editing task, without classifying whether the regions are semantically meant to be preserved or modified, which not only introduces potential background noises and wastes computation on non-edited regions. 

To address these two challenges explicitly, we propose \ourmtd, a training-free framework that efficiently skips redundant computation while preserving high fidelity and editing quality. \ourmtd consists of two parts;
(1) \textbf{SpotSelector} for detecting non-edited regions and skipping their computation, then reusing the original image information; 
(2) \textbf{Spotfusion} for providing temporally consistent context to regenerated tokens without advancing a full denoising process for the non-edited region.

\subsection{SpotSelector}
\label{sec:selector}

A central question in region-aware editing is determining which tokens truly require modification and which tokens should be preserved.  
SpotSelector addresses this problem by identifying non-edited regions early in the denoising process and routing them away from full DiT computation.  
By doing so, it enables \ourmtd to avoid unnecessary updates on stable regions while focusing computation on tokens that actually need to change.

To detect these non-edited regions, we leverage a key property of Rectified Flow: under its linear interpolation dynamics, the latent state at time $t$ admits a closed-form relation to the fully denoised latent state $X_0$.
\begin{equation}
\hat{X_{0}} = X_{t_i} - t_i \times\, v_\theta(X_{t_i}, C, t_i),
\quad i = T, \dots, 1
\label{eq_x0_reconstruction}
\end{equation}
And building on this reconstructed $\hat{X_{0}}$, we can decode it into an image.

As shown in Figure~\ref{fig:recon_sequence}, diffusion models exhibit a clear coarse to fine reconstruction pattern, where different spatial regions converge at different speeds.
Regions that rapidly become sharp and visually identical with the condition image at early timesteps have effectively \emph{stabilized}, which indicates them to be \emph{non-edited} areas.
In contrast, regions subject to editing continue to evolve throughout the denoising process under the given instruction.
A natural implication of this heterogeneous convergence behavior is that the model already exposes which regions are stable and which are still under refinement.

Building on this signal, Spotselector dynamically identifies and separates tokens corresponding to non-edited regions from those that require modification by comparing its early reconstruction $\hat{X}_0$ with the condition image latent $Y$. 
Tokens that are highly consistent with the condition image and stabilize early are classified as non-edit tokens, while the rest are treated as regenerated tokens. The latter continue through the full computation of DiT, while the former are routed away from full sampling updates and reuse the condition image token feature, thereby reducing unnecessary computation. This selective handling preserves non-edited region fidelity and yields substantial efficiency gains.
Although early stabilization provides a strong signal for identifying non-edited regions, efficiently and reliably selecting such tokens remains challenging, as distances in latent space do not necessarily reflect human-perceived similarity.
To obtain a similarity measure that better aligns with perceptual differences, we introduce a token-level perceptual score inspired by LPIPS\cite{zhang2018unreasonableeffectivenessdeepfeatures}, computed from VAE\cite{kingma2022autoencodingvariationalbayes} decoder activations.
Let $\phi_l(\cdot)$ denote the feature map extracted from decoder layer $l$.
The perceptual deviation of token $i$ is defined as
\begin{equation}
s_{\text{LPIPS}}(i)
=\sum_{l\in\mathcal{L}} w_l
\left|
\hat{\phi}_l(\hat{X}_0)_i-
\hat{\phi}_l(Y)_i
\right|_2^2,
\label{eq:lpips_like_token}
\end{equation}
where $w_l$ are layer weights and hats denote feature normalization.
This LPIPS-like formulation maps differences in early decoder features to a token-wise perceptual discrepancy that more faithfully reflects visually meaningful changes.

Having obtained a perceptual token score, we use it to decide whether each token should be actively denoised or directly reused from the reference.
Specifically, given a threshold $\tau$, we define the binary routing indicator
\begin{equation}
\begin{aligned}
r_{t,i} &= 1\!\left[s_{\text{LPIPS}}^{(t)}(i)\le\tau\right], \\[3pt]
\mathcal{R}_t &= \{ i : r_{t,i}=1 \}, \quad
\mathcal{A}_t = \{ i : r_{t,i}=0 \}.
\end{aligned}
\end{equation}
Tokens in the regenerated set $\mathcal{A}_t$ follow the full reverse-integration update (Eq.~\eqref{eq:inference}) to produce the desired edits.
In contrast, tokens in the non-edited set $\mathcal{R}_t$ are entirely removed from the DiT computation.
This design avoids redundant computation on non-edited regions.

At the final denoising step, we apply a lightweight latent consolidation.
Instead of propagating $\mathcal{R}_t$ through the network, all non-edited tokens are directly overwritten with their counterparts from the condition image latent before decoding into pixel space.
This guarantees that non-edited regions remain visually consistent with the condition image.

\subsection{SpotFusion}
\label{sec:token-fusion}

Routing non-edited tokens out of the DiT computation eliminates redundant updates but also removes their contextual contribution to regenerated regions.  
Because diffusion transformers rely heavily on cross-token attention to maintain spatial coherence, such naive removal can lead to noticeable degradation in edit quality.

A natural alternative is to cache and reuse the Key--Value (KV) pairs of non-edited tokens or reference-image tokens during attention.  
However, this approach introduces a temporal inconsistency: the cached KV pairs represent feature states frozen at the timestep when they were stored, while the hidden states of regenerated tokens continue to evolve throughout denoising.  

\begin{figure}[htbp]
    \centering
    \includegraphics[width=0.5\textwidth]{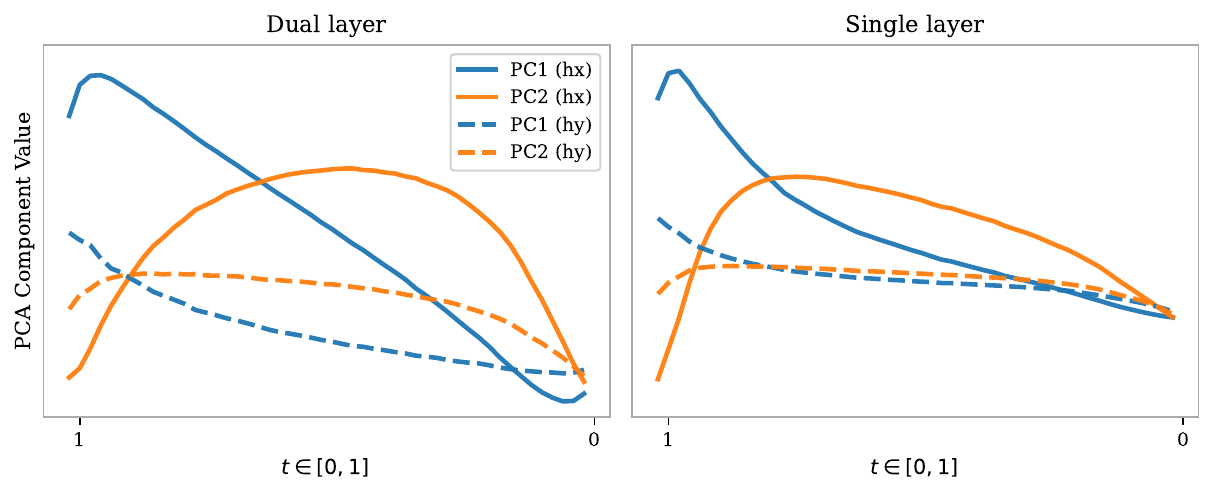}
    \caption{PCA trajectories of hidden states of non-edited tokens across dual stream layers and single stream layers. 
As denoising progresses, the trajectory of the generated image in the non-edited region ($x$) gradually approaches that of the condition image ($y$), 
indicating that their latent representations progressively align. 
Eventually, both trajectories overlap, suggesting that unedited regions converge to the same latent subspace, 
thereby maintaining strong background consistency and semantic preservation.}
    \label{fig:PCA}
\end{figure}

Unlike language transformers with static embeddings, DiTs maintain drift with timestep, causing cached KV pairs to become increasingly misaligned with the evolving edited-region features.  
This leads to temporal drift, context mismatch, and statistical discontinuity across layers.  
As demonstrated in our ablation study (Fig.~\ref{fig:ablation_fusion}), static KV caching results in clear degradation in editing fidelity.

To design a temporally consistent reuse mechanism, we first examine the temporal behavior of hidden states in non-edited regions.  
As shown in Fig.~\ref{fig:PCA}, the hidden representations of non-edited tokens ($x$-branch) and condition image tokens ($y$-branch) follow closely aligned trajectories in principal-component space:
\begin{enumerate}
  \item the reference branch exhibits strong temporal stability,
  \item non-edited tokens rapidly align with the reference trajectory after an initial transient phase,
  \item both $h_x^{(b,t)}$ and $h_y^{(b,t)}$ converge to the same latent subspace as $t\!\rightarrow\!0$.
\end{enumerate}
These observations suggest that non-edited regions evolve smoothly toward the condition image and therefore admit a consistent feature representation that can be gradually reinforced rather than statically cached.

\paragraph{SpotFusion}  
Motivated by this insight, we propose \textbf{SpotFusion}, a temporally consistent feature-reuse mechanism that blends cached non-edited features with reference-image features across timesteps.  
After the initial denoising steps, we cache the KV pairs of both the reference branch and the non-edited region tokens.  
These KV maps form \emph{condition cache} reused throughout the denoising process.

To avoid temporal mismatch with the evolving regenerated tokens, SpotFusion smoothly interpolates cached non-edited features toward reference features at every block and timestep.  
Let $\tilde{h}_x^{(b,t+1)}$ denote the cached hidden state at block $b$ from step $t{+}1$, and $h_y^{(b)}$ the corresponding reference feature.  
For non-edited tokens, the updated hidden state is:
\begin{equation}
\tilde{h}_x^{(b,t)}
= \alpha(t)\,\tilde{h}_x^{(b,t+1)}
+ (1-\alpha(t))\,h_y^{(b)},
\label{eq:fusion}
\end{equation}
where $\alpha(t)=\cos^2\!\left(\tfrac{\pi}{2}t\right)$ gradually shifts the representation from cached features to reference features as denoising proceeds.  
This interpolation is applied directly to the KV pairs:
\begin{align}
K^{(b)}_{t,i} &\leftarrow \alpha(t)K^{(b)}_{t+1,i} + (1-\alpha(t))K^{(b)}_{y,i}, \\
V^{(b)}_{t,i} &\leftarrow \alpha(t)V^{(b)}_{t+1,i} + (1-\alpha(t))V^{(b)}_{y,i}.
\end{align}

SpotFusion thus maintains coherent contextual signals for regenerated tokens, ensures temporal consistency in non-edited regions, and eliminates redundant computation associated with reprocessing the reference branch.  
Together, these advantages allow the model to preserve background fidelity while achieving substantial efficiency gains.

\paragraph{Partial attention calculation}
With SpotFusion providing temporally consistent KV caches for non-edited and condition image tokens, 
The DiT can perform attention using a small set of queries while retaining full spatial context.

During denoising, only the regenerated tokens require forward propagation through the DiT.  
Thus, the \textbf{Query} set is restricted to the regenerated tokens $\mathcal{A}_t$ together with the instruction--prompt tokens $P$:
\[
Q_{\text{active}} = [\,Q_P,\, Q_{\mathcal{A}_t}\,].
\]
Other tokens in non-edited regions and the condition image are skipped from computation.  
However, their contextual influence must remain available, so the Key--Value sets are completed by concatenating the cached features:
\[
K_{\text{full}} = [\,K_P,\, K_{\mathcal{A}_t},\, K_{\mathcal{R}_t}^{C},\, K_{Y}^{C}\,], 
V_{\text{full}} = [\,V_P,\, V_{\mathcal{A}_t},\, V_{\mathcal{R}_t}^{C},\,V_{Y}^{C}\,].
\]

Attention is then computed only on the active queries:
\begin{equation}
\mathrm{Attn}
=
\mathrm{softmax}\!\left(
    \frac{Q_{\text{active}}\, K_{\text{full}}^\top}{\sqrt{d}}
\right)
V_{\text{full}} .
\label{eq:partial_softmax}
\end{equation}

Here, the non-edited region and condition image token KV pairs $(K_{\mathcal{R}_t}^{C}, V_{\mathcal{R}_t}^{C})$ and $(K_Y^C, V_Y^C)$  
are retrieved directly from the SpotFusion condition cache.  

This partial-attention scheme concentrates computation precisely where edits occur,  
preserving contextual coherence through cached KV maps while avoiding redundant propagation through non-edited and reference regions.  

\begin{table*}[htbp]
    \centering
    \renewcommand{\arraystretch}{1.15}
    \resizebox{\textwidth}{!}{
    \begin{tabular}{l|ccccc|ccccc}
    \toprule
    \scriptsize
    \multirow{2}{*}{\textbf{Baselines}} &
    \multicolumn{5}{c}{\textbf{imgEdit-Benchmark}\cite{ye2025imgEdit}} &
    \multicolumn{5}{c}{\textbf{PIE-Bench++}\cite{huangParallelEditsEfficientMultiobject}} \\
    \cmidrule(lr){2-6} \cmidrule(lr){7-11}
     & CLIP$\uparrow$ & SSIMc$\uparrow$ & PSNR$\uparrow$ & DISTS$\downarrow$ & Speedup$\uparrow$ &
       CLIP$\uparrow$ & SSIMc$\uparrow$ & PSNR$\uparrow$ & DISTS$\downarrow$ & Speedup$\uparrow$ \\
    \midrule

    Original Inference &
    0.699 & 0.67 & 16.40 & 0.17 & 1.00$\times$ &
    0.741 & 0.791 & 18.76 & 0.136 & 1.00$\times$ \\
    \midrule

    FollowYourShape\cite{longFollowYourShapeShapeAwareImage2025} (single) &
    0.686 ($\downarrow$-0.013) & 0.47 ($\downarrow$-0.20) & 11.73 ($\downarrow$-4.67) & 0.27 ($\uparrow$0.10) & 0.33$\times$ ($\downarrow$-0.67) &
    0.714 ($\downarrow$-0.27) & 0.578 ($\downarrow$-0.213) & 12.91 ($\downarrow$-5.85) & 0.295 ($\uparrow$0.159) & 0.34$\times$ ($\downarrow$-0.66) \\

    FollowYourShape\cite{longFollowYourShapeShapeAwareImage2025} (multi) &
    0.688 ($\downarrow$-0.011) & 0.47 ($\downarrow$-0.20) & 11.47 ($\downarrow$-4.93) & 0.28 ($\uparrow$0.11) & 0.27$\times$ ($\downarrow$-0.73) &
    0.549 ($\downarrow$-0.192) & 0.583 ($\downarrow$-0.208) & 12.58 ($\downarrow$-6.18) & 0.298 ($\uparrow$0.162) & 0.28$\times$ ($\downarrow$-0.72) \\

    TeaCache\cite{liuTimestepEmbeddingTells2025} &
    \underline{0.698} ($\downarrow$0.001) & \underline{0.60} ($\downarrow$-0.07) & \underline{15.02} ($\downarrow$-1.38) & \underline{0.21} ($\uparrow$0.04) & \underline{3.43}$\times$ ($\uparrow$2.43) &
    0.735 ($\downarrow$-0.006) & \underline{0.764} ($\downarrow$-0.027) & \textbf{18.89} ($\uparrow$0.13) & \underline{0.144} ($\uparrow$0.08) & \underline{3.59}$\times$ ($\uparrow$2.59) \\

    TaylorSeer\cite{liuReusingForecastingAccelerating2025} &
    0.666 ($\downarrow$-0.033) & 0.52 ($\downarrow$-0.15) & 14.36 ($\downarrow$-2.04) & 0.37 ($\uparrow$0.20) & \textbf{3.61}$\times$ ($\uparrow$2.61) &
    \textbf{0.741} (0.00) & 0.755 ($\downarrow$-0.036) & 17.81 ($\downarrow$-0.95) & 0.159 ($\uparrow$0.023) & \textbf{3.86}$\times$ ($\uparrow$2.86) \\

    \rowcolor{gray!15}
    Ours &
    \textbf{0.699} (0.00) & \textbf{0.67} (0.00) & \textbf{16.45} ($\uparrow$0.05) & \textbf{0.16} ($\downarrow$-0.01) & 1.67$\times$ ($\uparrow$0.67) &
    \textbf{0.741} (0.00) & \textbf{0.792} ($\uparrow$0.01) & \underline{18.73} ($\downarrow$-0.03) & \textbf{0.136} (0.00) & 1.95$\times$ ($\uparrow$0.95) \\

    \bottomrule
    \end{tabular}}
    \caption{Comparison of models on \textbf{imgEdit-Benchmark} and \textbf{PIE-Bench++}. 
    }
    \label{tab:benchmark_speedup_delta}
\end{table*}

\section{Experiments}
\label{exp}

\subsection{Settings}

\paragraph{Implementation details}
All experiments are conducted on one single NVIDIA H200 GPU under the environment of CUDA~12.8 and PyTorch~2.9, scheduled with $T=50$ steps and a resolution of $1024\times1024$ at random seed 42.
\ourmtd employs a SpotSelector threshold $\tau=0.2$  of token saliency and a token-fusion weighting function $\alpha(t)=\cos^2(\pi t/2)$.
Condition image token features are cached after the initial stage at $t{=}4$ and reused during the remaining timesteps.

\paragraph{Datasets}
We evaluate \ourmtd on two image-editing benchmarks: PIE-Bench++~\cite{huangParallelEditsEfficientMultiobject}and imgEdit benchmark~\cite{ye2025imgEdit}.For PIE-Bench++, we select three subsets related to partial editing tasks: change object, add object, and delete object. For imgEdit benchmark, we focus on the single-turn setting and select the subsets most relevant to partial editing tasks, including adjust, background, extract, remove, replace, add, compose, and action. All images are standardized to $1024\times1024$ resolution.

\begin{table*}[htbp]
    \centering
    \renewcommand{\arraystretch}{1.15}
    \resizebox{\textwidth}{!}{
    \begin{tabular}{l|ccccccccc}
    \toprule
    \scriptsize
    \textbf{Baselines} & Adjust & Background & Extract & Remove & Replace & Add & Compose & Action & Average\\
    \midrule
    Original Inference & 
    3.73 & 3.64 & 4.09 & 4.80 & 4.46 & 3.78 & 2.75 & 4.01 & 3.91 \\
    \midrule

    FollowYourShape\cite{longFollowYourShapeShapeAwareImage2025} { (single)} &
    3.27 ($\downarrow$-0.46) &
    3.27 ($\downarrow$-0.37) &
    3.10 ($\downarrow$-0.99) &
    4.67 ($\downarrow$-0.13) &
    4.10 ($\downarrow$-0.36) &
    3.16 ($\downarrow$-0.62) &
    2.32 ($\downarrow$-0.43) &
    3.54 ($\downarrow$-0.47) &
    3.43 ($\downarrow$-0.48) \\

    FollowYourShape\cite{longFollowYourShapeShapeAwareImage2025} { (multi)} &
    3.15 ($\downarrow$-0.58) &
    3.31 ($\downarrow$-0.33) &
    3.13 ($\downarrow$-0.96) &
    \textbf{4.77} ($\downarrow$-0.03) &
    4.00 ($\downarrow$-0.46) &
    3.18 ($\downarrow$-0.60) &
    2.38 ($\downarrow$-0.37) &
    3.36 ($\downarrow$-0.65) &
    3.41 ($\downarrow$-0.50) \\

    TaylorSeer\cite{liuReusingForecastingAccelerating2025} &
    3.51 ($\downarrow$-0.22) &
    3.39 ($\downarrow$-0.25) &
    3.56 ($\downarrow$-0.53) &
    4.66 ($\downarrow$-0.14) &
    \underline{4.19} ($\downarrow$-0.27) &
    \textbf{3.66} ($\downarrow$-0.12) &
    2.41 ($\downarrow$-0.34) &
    3.96 ($\downarrow$-0.05) &
    3.67 ($\downarrow$-0.24) \\

    TeaCache\cite{liuTimestepEmbeddingTells2025} &
    \underline{3.55} ($\downarrow$-0.18) &
    \underline{3.41} ($\downarrow$-0.23) &
    \textbf{3.65} ($\downarrow$-0.44) &
    4.64 ($\downarrow$-0.16) &
    4.19 ($\downarrow$-0.27) &
    3.51 ($\downarrow$-0.27) &
    \underline{2.51} ($\downarrow$-0.24) &
    \underline{4.17} ($\uparrow$+0.16) &
    \underline{3.70} ($\downarrow$-0.21) \\

    \rowcolor{gray!15}
    Ours &
    \textbf{3.57} ($\downarrow$-0.16) &
    \textbf{3.43} ($\downarrow$-0.21) &
    \underline{3.60} ($\downarrow$-0.49) &
    \underline{4.72} ($\downarrow$-0.08) &
    \textbf{4.41} ($\downarrow$-0.05) &
    \underline{3.64} ($\downarrow$-0.14) &
    \textbf{2.65} ($\downarrow$-0.10) &
    \textbf{4.14} ($\uparrow$+0.13) &
    \textbf{3.77} ($\downarrow$-0.14) \\
    \bottomrule
    \end{tabular}}
    \caption{VL score comparison of eight subsets of \textbf{imgEdit-Benchmark}.}
    \label{tab:VL_comparison}
\end{table*}

\begin{figure*}[t]
    \centering
    \includegraphics[width=\textwidth]{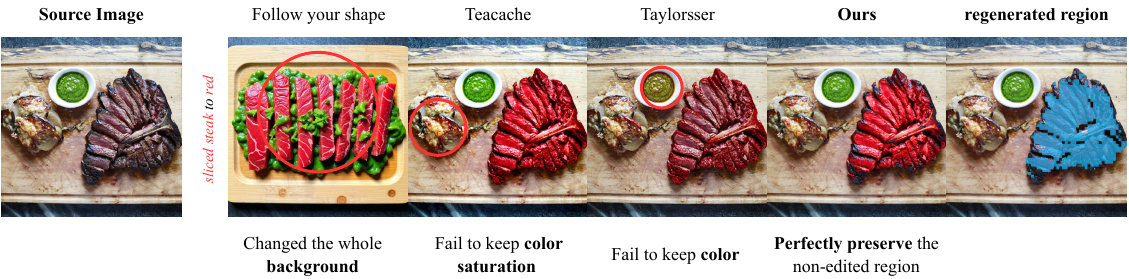}
    \caption{Non-edited preservation comparison across different models. Prior methods either modify unnecessary background regions or distort color consistency, whereas our method preserves non-edited areas faithfully while applying accurate edits.}
    \label{fig:model_comparison}
\end{figure*}

\paragraph{Evaluation metrics}
We construct a comprehensive evaluation protocol that jointly measures editing quality and computational efficiency.
For quality assessment, we adopt four complementary criteria—semantic alignment, structural consistency, perceptual fidelity, and overall editing accuracy.
Semantic alignment between the edited results and textual instructions is evaluated using CLIP similarity~\cite{radford2021learningtransferablevisualmodels}, while PSNR, SSIM, and DISTS quantify structural preservation and perceptual fidelity.
Together, these metrics capture how well the edits follow the intended instructions while maintaining consistency in non-edited regions.
For efficiency assessment, we additionally report the average inference latency and the relative speedup.

\paragraph{Baselines}
We compare \ourmtd against a diverse set of representative baselines spanning both cache acceleration and percise editing paradigms.
We include cache accelerating methods such as TaylorSeer~\cite{liuReusingForecastingAccelerating2025} and TeaCache~\cite{liuTimestepEmbeddingTells2025}, which improve diffusion inference efficiency through feature reuse, forecasting, or adaptive caching strategies. Also, we evaluate precise-editing approaches such as Follow-Your-Shape~\cite{longFollowYourShapeShapeAwareImage2025}, which emphasize spatially accurate, mask-free editing by explicitly preserving structural cues during generation. Specifically, we evaluate FollowYourShape under two ControlNet configurations. The “single” setting denotes that the model is driven by a single ControlNet, where we adopt the depth ControlNet as in the original implementation. The “multi” setting instead employs a multi-ControlNet composition, where multiple structural guidance branches are jointly used to provide richer geometric and semantic constraints.

\subsection{Results analyse}

\paragraph{Quantitative results}
Table~\ref{tab:benchmark_speedup_delta} summarizes the results on imgEdit-Benchmark and PIE-Bench++.SpotEdit achieves a highly favorable trade-off between editing fidelity and computational efficiency by explicitly identifying and skipping non-edited regions during denoising.
On IMGedit-Benchmark, SpotEdit matches or slightly surpasses the original inference results, while achieving a 1.67× speedup.
In contrast, cache-based methods such as TaylorSeer obtain higher acceleration (3.61×) but suffer from clear all degradation, while precise-editing methods like FollowYourShape introduce severe distortion to non-edited areas.
Similar trends hold on PIE-Bench++, where SpotEdit maintains 18.73 PSNR and 0.792 SSIMc, outperforming all baselines in quality while still reaching 1.95× speedup.

As vision-language (VL) scores across eight instruction types on the IMGedit-Benchmark (Table~\ref{tab:VL_comparison}) illustrated, SpotEdit achieves the highest VL score (3.77) among all methods, with strong performance on complex instructions like Replace (4.41) and Compose (2.65), and only a slight drop (-0.14) from original inference. This demonstrates that SpotEdit effectively completes the intended edits while preserving structural consistency.
These results show that SpotEdit allocates computation only to regions requiring change, preserving the structure and consistency of unedited areas, validating our core principle—edit what needs to be edited
\paragraph{Qualitative results}
Figure~\ref{fig:model_comparison} shows baseline methods struggle to keep the background intact. Our method preserves non-edited regions almost perfectly while delivering the desired edit.
Figure~\ref{fig:Qualitative_Results} demonstrates that our method works well across diverse editing instructions, and only regenerates the needed region. 

\subsection{Ablation study}

We conduct ablation studies on three key components: \emph{Token Fusion}, \emph{Condition Cache}, and \emph{Reset mechanism}.

\begin{figure}[htbp]
    \centering
    \includegraphics[width=0.5\textwidth]{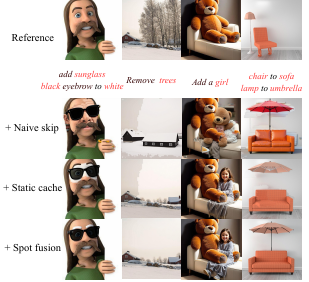}
    \caption{ The qualitative ablation study on token fusion}
    \label{fig:ablation_fusion}
\end{figure}

\paragraph{Token fusion} We compare Spot Fusion with two variants:
(1) \textit{Naive Skip}, which directly discards non-edited tokens without caching, leading to loss of context for regenerated regions;
(2) \textit{Static Token Fusion}, which reuses cached unedited tokens without aligning them with the condition image.
As shown in Figure~\ref{fig:ablation_fusion}, both variants introduce artifacts or inconsistency. In contrast, Spot Fusion preserves background fidelity and edit quality, highlighting the necessity of adaptive token-level blending.

\begin{table}[h!]
\centering

\label{tab:ablation_reset}
\resizebox{0.47\textwidth}{!}{
\begin{tabular}{l|ccccc}

\toprule
Method & CLIP$\uparrow$ & SSIM$_c$$\uparrow$ & PSNR$\uparrow$ & DISTS$\downarrow$ & Speedup$\uparrow$ \\
\midrule
Default & \textbf{0.741} & \textbf{0.792} & \textbf{18.730} & \textbf{0.136} & 1.95 \\
w/o Reset & 0.738 & 0.782 & 17.10 & 0.154 & \textbf{2.25} \\
\bottomrule
\end{tabular}}
\caption{Ablation on \textbf{Reset} mechanism.}
\end{table}
\begin{table*}[t!]
    \centering
    \renewcommand{\arraystretch}{1.15}
    \resizebox{\textwidth}{!}{
    \begin{tabular}{l|ccccc|ccccc}
    \toprule
    \scriptsize
    \multirow{2}{*}{\textbf{Baselines}} &
    \multicolumn{5}{c}{\textbf{imgEdit-Benchmark}\cite{ye2025imgEdit}} &
    \multicolumn{5}{c}{\textbf{PIE-Bench++}\cite{huangParallelEditsEfficientMultiobject}} \\
    \cmidrule(lr){2-6} \cmidrule(lr){7-11}
     & CLIP$\uparrow$ & SSIMc$\uparrow$ & PSNR$\uparrow$ & DISTS$\downarrow$ & Speedup$\uparrow$ &
       CLIP$\uparrow$ & SSIMc$\uparrow$ & PSNR$\uparrow$ & DISTS$\downarrow$ & Speedup$\uparrow$ \\
    \midrule

    Qwen-Image-Edit &
    \textbf{0.688} & 0.522 & 14.23 & 0.22 & 1.00$\times$ &
    \textbf{0.724} & 0.74 & 18.32 & 0.16 & 1.00$\times$ \\

    Ours &
    0.686 ($\uparrow$0.002) & \textbf{0.524} ($\uparrow$0.02) & \textbf{14.24} ($\uparrow$0.01) & \textbf{0.21} ($\downarrow$-0.01) & \textbf{1.59}$\times$ ($\uparrow$0.59) &
    0.722 ($\downarrow$0.002) & \textbf{0.77} ($\uparrow$0.03) & \textbf{19.40} ($\uparrow$1.08) & \textbf{0.15} ($\downarrow$-0.01) & \textbf{1.72}$\times$ ($\uparrow$0.72) \\

    \bottomrule
    \end{tabular}
    }
    \caption{Comparison between Qwen-Image-Edit and SpotEdit on \textbf{imgEdit-Benchmark} and \textbf{PIE-Bench++}.}
    \label{tab:benchmark_speedup_delta_qwen_ours}
\end{table*}
\vspace{-1em}

\paragraph{Condition cache}
We compare SpotEdit’s full condition caching with a variant that only caches non-edited tokens, while recomputing condition image features at each timestep.
Our approach caches both condition image and unedited region features, enabling greater reuse and faster inference.
Though the uncached variant achieves slightly better PSNR (19.15 vs. 18.73), it is notably slower (1.24× vs. 1.95×).
This shows that caching the condition image contributes significantly to generation efficiency without harming visual quality.

\paragraph{Reset mechanism}
To further ensure numerical stability, we introduce a periodic reset mechanism that forces cached tokens to refresh after an interval. 
This prevents the accumulation of numerical errors across timesteps.
Without Reset, the model gains a slight speedup (from 1.95$\times$ to 2.25$\times$) but suffers a clear quality drop: 
PSNR decreases by 1.6 dB, and DISTS increases by 0.018. 
This suggests that cumulative cache error gradually deviates unedited regions from their reference alignment. 
Periodic reset incurs negligible overhead yet effectively stabilizes the denoising trajectory, ensuring consistent structure preservation across steps.

\begin{table}[t!]
\centering

\label{tab:ablation_cache}
\resizebox{0.47\textwidth}{!}{
\begin{tabular}{l|ccccc}
\toprule
Method & CLIP$\uparrow$ & SSIM$_c$$\uparrow$ & PSNR$\uparrow$ & DISTS$\downarrow$ & Speedup$\uparrow$ \\
\midrule
Defualt & 0.741 & 0.792 & 18.730 & 0.136 & \textbf{1.95} \\
w/o Condition Cache & \textbf{0.787} & \textbf{0.801} & \textbf{19.155} & \textbf{0.131} & 1.24 \\
\bottomrule                                           
\end{tabular}}
\caption{Ablation on \textbf{Condition Cache}.}
\end{table}

\subsection{Additional results on Qwen-Image-Edit}
As shown in Table~\ref{tab:benchmark_speedup_delta_qwen_ours}, when SpotEdit is applied to Qwen-Image-Edit\cite{wuQwenImageTechnicalReport2025}, SpotEdit preserves the original model’s background fidelity perfectly, showing only +0.01 PSNR and -0.01 DISTS difference on imgEdit, while achieving 1.59$\times$ acceleration, and on PIE-Bench, it even improves fidelity (+0.03 SSIMc, +1.08 PSNR, -0.01 DISTS) with 1.72$\times$ acceleration.

\section{Conclusion}
\label{sec:conclusion}

In this paper, we present {SpotEdit}, a training-free, region-aware framework that accelerates instruction-based image editing by following a simple principle: \textit{edit what needs to be edited}.
Unlike prior methods that denoise all image tokens uniformly, SpotEdit identifies and isolates edited regions at the token level, avoiding unnecessary computation on unedited areas.
With {SpotSelector}, we indentify and remove non-edited tokens from processing and directly restore them from the condition image.
{SpotFusion} further refines this with consistent feature reuse, ensuring contextual alignment.
Extensive experiments show that SpotEdit delivers substantial speedups while preserving editing precision and background consistency.

\clearpage

\appendix

\maketitlesupplementary
\renewcommand{\thesection}{\Alph{section}}

\setcounter{table}{0}
\renewcommand{\thetable}{S\arabic{table}}

\setcounter{figure}{0}
\renewcommand{\thefigure}{S\arabic{figure}}

\setcounter{algocf}{0}
\renewcommand{\thealgocf}{S\arabic{algocf}}

\section{Pseudocode of SpotEdit}

\begin{algorithm}[h]
\caption{SpotEdit: Selective Region Editing with Diffusion Transformers}
\label{alg:spotedit}
\SetKwInOut{KwInput}{Input}
\SetKwInOut{KwOutput}{Output}
\KwInput{Diffusion Transformer $\Phi$, Editing Instruction $P$,
Condition Image Latent $Y$,
Initial Noise $X_T$, Total Steps $T$, Initial Stage Steps $K_{init}$, Threshold $\tau$ ,Time Schedule $\{t_i\}_{i=0}^{T}$ where $t_T =1, t_0=0$}

\KwOutput{Edited Image $Img$}

Initialize Condition Cache $(K_Y, V_Y)$;\\
Initialize Feature Cache $(K_{cache}, V_{cache}) \leftarrow \emptyset$;\\

\For{$i \leftarrow T, T-1, \dots, T-K_{init} + 1$}{
    $v_{t_i}, (K_{curr}, V_{curr}) \leftarrow \Phi(X_{t_i}, t_i, P, Y)$;
    
    $X_{t_{i-1}} \leftarrow X_{t_i} - (t_{i} - (t_{i-1})) \cdot v_{t_i}$;\\
    
    $\hat{X}^{t_i}_{0} \leftarrow X_{t_i} - t_i \cdot v_{t_i}$;
    
    $(K_{cache}, V_{cache}) \leftarrow (K_{curr}, V_{curr})$;
}

\For{$i \leftarrow T-K_{init}, \dots, 1$}
{
  $[ \mathcal{A}_{t_i},\mathcal{R}_{t_i}]\leftarrow \textbf{SpotSelector}(\hat{X}^{t_i}_0, Y, \tau)$;
    
    \For{each transformer block $b$}{
        $K^{(b)}_{\mathcal{R}_{t_i}} \leftarrow \alpha(t_i) K^{(b)}_{\mathcal{R}_{t_{i+1}}} + (1-\alpha(t_i)) K^{(b)}_{Y}$;\\
        $V^{(b)}_{\mathcal{R}_{t_i}} \leftarrow \alpha(t_i) V^{(b)}_{\mathcal{R}_{t_{i+1}}} + (1-\alpha(t_i)) V^{(b)}_{Y}$;
    }
    
    Construct Queries $Q_{\text{active}}$ for tokens $\in \mathcal{A}_{t_i}$\\
    $Q_{\text{active}} \leftarrow [Q_P,Q_{\mathcal{A}_{t_i}}]$\\
    Construct Keys and Values \\
    $K_{\text{full}} \leftarrow [K_P,K_{\mathcal{A}_{t_i}}, K_{\mathcal{R}_{t_i}},K_Y]$\\
    $V_{\text{full}} \leftarrow [V_P,V_{\mathcal{A}_{t_i}}, V_{\mathcal{R}_{t_i}},V_Y]$;\\
    
    $v_{t_i}[\mathcal{A}_{t_i}] \leftarrow \text{Attention}(Q_{\text{active}}, K_{\text{full}}, V_{\text{full}})$;
    $X_{t_{i-1}}[\mathcal{A}_{t_i}] \leftarrow X_t[\mathcal{A}_{t_i}] - (t_i - t_{i-1}) \cdot v_{t_i}[\mathcal{A}_{t_i}]$;\\
    
    $\hat{X}^{t_i}_{0} \leftarrow X_{t_{i}}-t_{i}\cdot v_{t_i}$
    
}

Identify final non-edited regions $$\mathcal{R}_{final}$$

$X_0^{final}[\mathcal{A}_{final}] \leftarrow X_0[\mathcal{A}_{final}]$;\\

$X_0^{final}[\mathcal{R}_{final}] \leftarrow Y[\mathcal{R}_{final}]$;

$Img \leftarrow \text{VAE}(X_0^{final})$;

\Return $Img$
\end{algorithm}

\begin{algorithm}[t]
\caption{SpotSelector: LPIPS-like Perceptual Scoring}
\label{alg:spotselector_metric}
\SetKwInOut{KwInput}{Input}
\SetKwInOut{KwOutput}{Output}

\KwInput{
    Reconstructed latent $\hat{X}_0 \in \mathbb{R}^{N \times C}$, 
    Conditional image latent $Y \in \mathbb{R}^{N \times C}$, \\
    VAE Decoder shallow layers $\mathcal{L}$, \\Spatial dimensions $(H, W)$, Patch size $p$
}
\KwOutput{Regenerate region and non-edited region indices $[\mathcal{A}_{t_i},\mathcal{R}_{t_i}]$}

$F_{\hat{x}} \leftarrow \{ \phi_l(\hat{x}_{input}) \mid l \in \mathcal{L} \}$; \\
$F_{y} \leftarrow \{ \phi_l(y_{input}) \mid l \in \mathcal{L} \}$\;

\BlankLine
Initialize spatial score map $M \leftarrow \mathbf{0}$\;
\For{each layer $l \in \mathcal{L}$}{
    $D_l \leftarrow \| \text{Norm}(F_{\hat{x}}^{(l)}) - \text{Norm}(F_{y}^{(l)}) \|_2^2$\;
    $D_l^{aligned} \leftarrow \text{Resize}(D_l, \text{size}\leftarrow(H, W))$\;
    $M \leftarrow M + D_l^{aligned}$\;
}
$M \leftarrow M / |\mathcal{L}|$

$S_{pooled} \leftarrow \text{AvgPool}(M, \text{kernel}, \text{stride})$\;
$S_{token} \leftarrow \text{Flatten}(S_{pooled})$\;
$[\mathcal{A}_{t_i},\mathcal{R}_{t_i}] \leftarrow S_{token} < \tau$

\Return $[\mathcal{A}_{t_i},\mathcal{R}_{t_i}]$
\end{algorithm}

\newpage

\section{Compatibility with Existing Acceleration Methods}
\label{sec:compatibility}

A key advantage of \textbf{SpotEdit} is that it is \textbf{orthogonal} to existing acceleration techniques for Diffusion Transformers (DiTs).  
While prior methods accelerate along the 
\emph{temporal}, \emph{feature}, or \emph{attention} dimensions, SpotEdit accelerates computation along the 
\emph{spatial} by skipping non-edited regions. Importantly, the acceleration dimensions targeted by these methods are inherently complementary to 
the spatial acceleration pursued by SpotEdit. Rather than competing with SpotEdit’s region spotting and token skipping mechanism, they operate along orthogonal axes, enabling their effects to be combined additively for further speed improvements.

\subsection{General compatibility with full-token computation accelerators}

Let the set of regenerated tokens selected by SpotSelector be $\mathcal{A}$ and non-edited tokens be $\mathcal{R}$.

SpotFusion reconstructs $(K,V)$ values for computation of all regenerated tokens 
$\mathcal{A}$:
\[
    K_{\text{full}} =[K_P,K_{\mathcal{A}}, K_{\mathcal{R}},K_Y]\\
    V_{\text{full}} = [V_P,V_{\mathcal{A}}, V_{\mathcal{R}},V_Y]
\]

This reconstruction ensures that the edited region forms a 
\textbf{closed and condition-complete subgraph} of the DiT model.  
Thus, any acceleration operator $\mathcal{O}_{acc}$ that assumes full attention context may be applied to the edited region alone:
\[
\mathcal{O}_{acc}(X_{\mathcal{A}})
\;\oplus\;
X_{\mathcal{R}}^{\text{cache}},
\]
where $\oplus$ denotes spatial concatenation.

Since SpotEdit does not change the functional form of the DiT computation and only restricts its spatial domain, it is fully compatible with other temporal accelerators.

\subsection{Compatibility with TeaCache and TaylorSeer}

To further demonstrate that \ourmtd is orthogonal to temporal accelerators, we integrate  
\textbf{TeaCache} and \textbf{TaylorSeer} into the SpotEdit pipeline.  
In both cases, the edited region subgraph produced by SpotSelector and reconstructed by SpotFusion
forms a self-contained full-token region, ensuring that existing reuse-based accelerators operate without modification.

Formally, for any accelerator $\mathcal{O}_{acc}$ applied on the edited tokens, the composite update takes the unified form:
\[
\mathcal{F}_{\text{SpotEdit}+\mathcal{O}_{acc}}(X)
\leftarrow
\mathcal{O}_{acc}\big(X_{\mathcal{A}})
\;\oplus\;
X_{\mathcal{R}}^{\text{cache}},
\]
where the cached non-edited tokens remain valid due to SpotFusion's full reconstruction of $(K,V)$.

\paragraph{TeaCache}
TeaCache performs timestep feature reuse through caching.  
Since SpotFusion regenerates complete attention states for edited tokens, TeaCache cached residuals remain fully compatible and can be directly reused inside the edited subgraph.

\paragraph{TaylorSeer}
TaylorSeer approximates residuals via local Taylor-series predictions.  
Because the edited subgraph satisfies the same continuous-time latent dynamics, the Taylor approximation computed on $X_{\mathcal{A}}$ remains valid, while non-edited tokens continue using cached features.

\subsection{Experimental results}

Table~\ref{tab:imgedit_mixture_results} and Table~\ref{tab:PIE-mixture_results} report speed and quality metrics for SpotEdit combined with TeaCache and TaylorSeer.  
Both integrations remain stable and improve efficiency while preserving editing quality.

These results empirically confirm TeaCache and TaylorSeer integrate seamlessly into SpotEdit.

\begin{table}[htbp]
    \centering
    \renewcommand{\arraystretch}{1.15}
    \resizebox{\columnwidth}{!}{
    \begin{tabular}{l|ccccc}
    \toprule
    \scriptsize
    \textbf{Baselines} &
    CLIP$\uparrow$ & SSIMc$\uparrow$ & PSNR$\uparrow$ &
    DISTS$\downarrow$ & Speedup$\uparrow$ \\
    \midrule

    Original Inference &
    0.699 & 0.67 & 16.40 & 0.17 & 1.00$\times$ \\
    \midrule

    TeaCache\cite{liuTimestepEmbeddingTells2025} &
    \textbf{0.698} ($\downarrow$0.001) &
    0.60 ($\downarrow$0.07) &
    15.02 ($\downarrow$1.38) &
    0.21 ($\uparrow$0.04) &
    3.43$\times$ ($\uparrow$2.43) \\

    TaylorSeer\cite{liuReusingForecastingAccelerating2025} &
    0.666 ($\downarrow$0.033) &
    0.52 ($\downarrow$0.15) &
    14.36 ($\downarrow$2.04) &
    0.37 ($\uparrow$0.20) &
    3.61$\times$ ($\uparrow$2.61) \\
    \midrule

    SpotEdit--TeaCache &
    0.695 ($\downarrow$0.004) &
    \textbf{0.62} ($\downarrow$0.05) &
    \textbf{15.57} ($\downarrow$0.83) &
    \textbf{0.19} ($\uparrow$0.02) &
    \textbf{3.94}$\times$ ($\uparrow$2.94) \\

    SpotEdit--TaylorSeer &
    \underline{0.698} ($\downarrow$0.001) &
    \underline{0.61} ($\downarrow$0.06) &
    \underline{15.50} ($\downarrow$0.90) &
    \underline{0.19} ($\uparrow$0.02) &
    \underline{3.85}$\times$ ($\uparrow$2.85) \\

    \bottomrule
    \end{tabular}}
    \caption{Comparison of models on \textbf{imgEdit-Benchmark}.}
    \label{tab:imgedit_mixture_results}
\end{table}

\begin{table}[htbp]
    \centering
    \renewcommand{\arraystretch}{1.15}
    \resizebox{\columnwidth}{!}{
    \begin{tabular}{l|ccccc}
    \toprule
    \scriptsize
    \textbf{Baselines} &
    CLIP$\uparrow$ & SSIMc$\uparrow$ & PSNR$\uparrow$ &
    DISTS$\downarrow$ & Speedup$\uparrow$ \\
    \midrule

    Original Inference &
    0.741 & 0.791 & 18.76 & 0.136 & 1.00$\times$ \\
    \midrule

    TeaCache\cite{liuTimestepEmbeddingTells2025} &
    0.735 ($\downarrow$0.006) &
    0.764 ($\downarrow$0.027) &
    \underline{18.89} ($\uparrow$0.13) &
    0.144 ($\uparrow$0.008) &
    3.59$\times$ ($\uparrow$2.59) \\

    TaylorSeer\cite{liuReusingForecastingAccelerating2025} &
    \underline{0.741} (0.00) &
    0.755 ($\downarrow$0.036) &
    17.81 ($\downarrow$0.95) &
    0.159 ($\uparrow$0.023) &
    3.86$\times$ ($\uparrow$2.86) \\
    \midrule

    SpotEdit--TeaCache &
    0.740 ($\downarrow$0.0005) &
    \textbf{0.797} ($\uparrow$0.006) &
    \textbf{18.98} ($\uparrow$0.22) &
    \textbf{0.133} ($\downarrow$0.003) &
    \textbf{4.28}$\times$ ($\uparrow$3.28) \\

    SpotEdit--TaylorSeer &
    \textbf{0.743} ($\uparrow$0.002) &
    \underline{0.783} ($\downarrow$0.008) &
    18.50 ($\downarrow$0.26) &
    \underline{0.142} ($\uparrow$0.006) &
    \underline{4.16}$\times$ ($\uparrow$3.16) \\

    \bottomrule
    \end{tabular}}
    \caption{Comparison of models on \textbf{PIE-Bench++}.}
    \label{tab:PIE-mixture_results}
\end{table}

\newpage

\section{Discussion between $\ell_2$-distance and LPIPS-like score}

We further justify the use of the LPIPS-like score in \textbf{SpotSelector} by analyzing the spectral bias inherent to different similarity metrics.
\begin{figure}[htbp]
    \centering
    \includegraphics[width=0.5\textwidth]{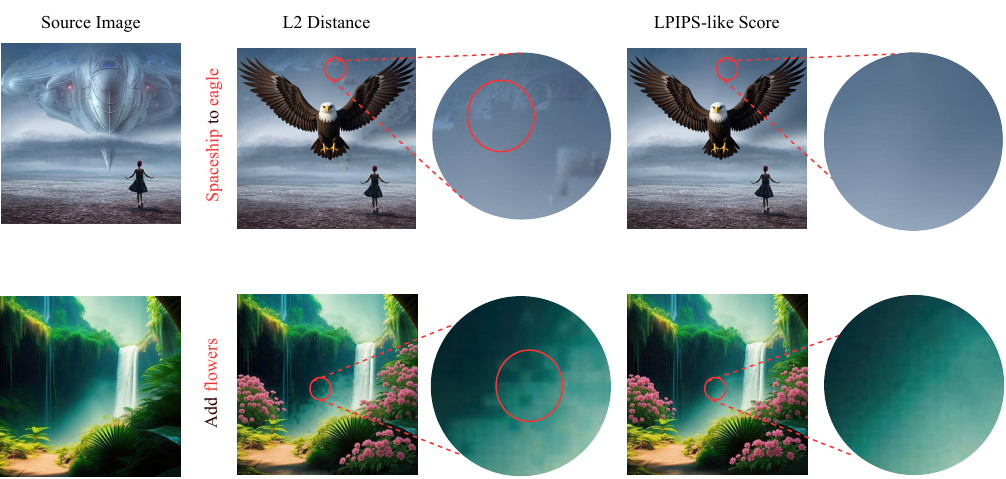}
    \caption{\textbf{$\ell_2$ Distance vs. LPIPS-like Score.} 
    Low-frequency changes (e.g., brightness) produce overly large $\ell_2$ responses, while subtle high-frequency texture edits barely affect it. 
    As shown in the first sample, $\ell_2$ incorrectly preserves the spaceship that should have been removed; in the second sample, it misclassifies background tokens as regenerate tokens, causing unnecessary regeneration and background degradation. 
    LPIPS-like features avoid these failures by operating in a perceptually aligned feature space.}
    \label{fig:L2-LPIPS-comparison}
\end{figure}

\textbf{Limitations of latent $\ell_2$ distance.}
Latent representations in diffusion models are highly compressed. A point-wise $\ell_2$ distance is dominated by low-frequency components such as global brightness and color statistics. As noted by Zhang et al.~\cite{zhang2018unreasonableeffectivenessdeepfeatures}, pixel-wise metrics assume independence across dimensions and thus fail to capture structural degradations like blur, which primarily remove high-frequency content but induce only mild $\ell_2$ deviations.  
In selective editing, this bias produces two failure modes:  
(1) low-frequency shifts, such as brightness changes, disproportionately inflate $\ell_2$ and falsely mark a region as `regenerated', and  
(2) high-frequency feature changes with similar low-frequency features remain undetected, causing truly edited regions to be misclassified as `non-edited'.  

\textbf{Perceptual transferability via the VAE decoder.}
Although the observations of Zhang et al.~were established in the pixel domain, their core principle, deep features reflect perceptual similarity better than raw vectors extends naturally to latent diffusion models.  
The VAE decoder provides a non-linear mapping from compressed latent space back to perceptual image manifolds. Its intermediate activation maps recover spatial structure and high-frequency cues that are not explicitly represented in raw latents.  

Building on this insight, our LPIPS-like score measures distances between decoder features, analogous to evaluating perceptual differences through VGG features in standard LPIPS. This grants two advantages:

It captures high-frequency discrepancies essential for determining whether a region was genuinely edited.  
And it ensures alignment between a non-edited region $\mathcal{R}_t$ and the condition image $Y$ not only in coarse color tone but also in fine-grained spatial patterns.

By leveraging decoder deep features, LPIPS-like score provides a perceptually faithful descriptor for region stability and editing consistency. This resolves the spectral bias of $\ell_2$ and enables robust token selection in SpotSelector.
\section{More Visualization Results}

\begin{figure*}[t]
    \centering
    \includegraphics[width = \textwidth]{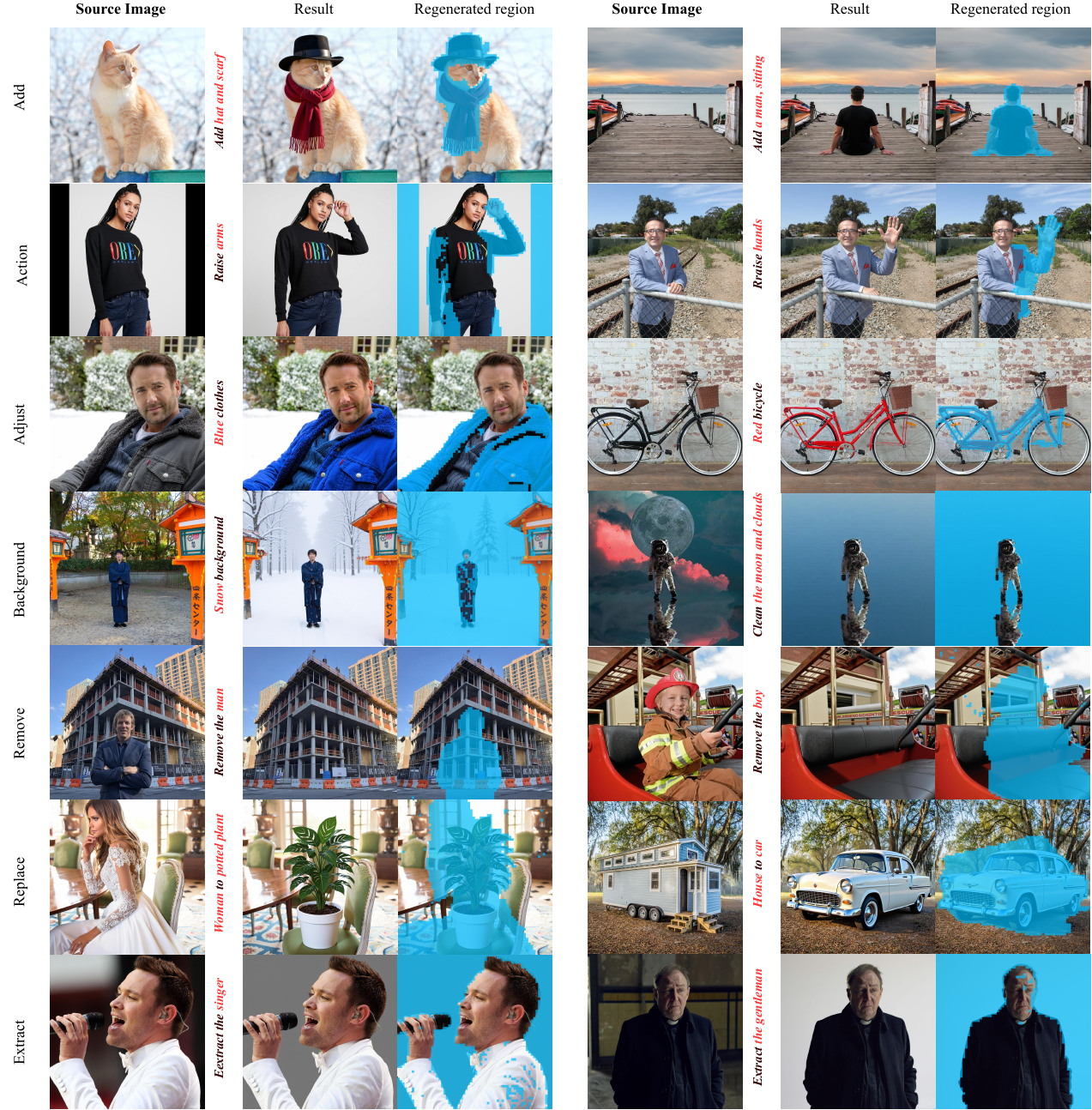}
    \caption{More results of SpotEdit with various editing tasks}
    \label{fig:add_results}
\end{figure*}

As shown in Fig.~\ref{fig:add_results}, we provide additional qualitative results of SpotEdit across a wide range of instruction-based editing tasks, including \textit{add}, \textit{action}, \textit{adjust}, \textit{background}, \textit{remove}, \textit{replace}, and \textit{extract}. These examples further demonstrate the versatility and robustness of our framework in handling diverse semantic manipulations while maintaining high fidelity in non-edited regions.

\clearpage
{
    \small
    \bibliographystyle{ieeenat_fullname}
    \bibliography{main}
}


\end{document}